\pdfoutput=1

\documentclass[11pt]{article}

\usepackage{naacl2021}
\usepackage{pdfpages}

\usepackage{times}
\usepackage{latexsym}

\usepackage[T1]{fontenc}
\usepackage[utf8]{inputenc}
\usepackage[ruled,noend,linesnumbered]{algorithm2e}
\usepackage{diagbox}
\usepackage{colortbl}
\usepackage{subcaption}
\usepackage{floatrow}
\usepackage{xcolor}
\usepackage{graphicx}
\usepackage{microtype}

\title{BBAEG: Towards BERT-based Biomedical Adversarial Example Generation for Text Classification}

\author{Ishani Mondal \\
  Microsoft Research India\\
  \texttt{ishani340@gmail.com}}

\begin{document}
\maketitle
\begin{abstract}
Healthcare predictive analytics aids medical decision-making, diagnosis prediction and drug review analysis. Therefore, prediction accuracy is an important criteria which also necessitates robust predictive language models. However, the models using deep learning have been proven vulnerable towards insignificantly perturbed input instances which are less likely to be misclassified by humans. Recent efforts of generating adversaries using rule-based synonyms and BERT-MLMs have been witnessed in general domain, but the ever-increasing biomedical literature poses unique challenges. We propose \textbf{BBAEG} (Biomedical BERT-based Adversarial Example Generation), a black-box attack algorithm for biomedical text classification, leveraging the strengths of both domain-specific synonym replacement for biomedical named entities and BERT-MLM predictions, spelling variation and number replacement. Through automatic and human evaluation on two datasets, we demonstrate that BBAEG performs stronger attack with better language fluency, semantic coherence as compared to prior work.
\end{abstract}

\section{Introduction}
Recent studies have exposed the importance of biomedical NLP in the well-being of human-beings, analyzing the critical process of medical decision-making. However, the dialogue managing tools  targeted for medical conversations \cite{Zhang2020MIEAM}, \cite{campillos-llanos-etal-2017-automatic}, \cite{kazi-kahanda-2019-automatically} between patients and healthcare providers in assisting diagnosis may generate certain insignificant perturbations (spelling errors, paraphrasing), which when fed to the classifier to determine the type of diagnosis required/detecting adverse drug effects/drug recommendation, might provide unreasonable performance. 
Insignificant perturbations might also creep in from the casual language expressed in the tweets \cite{Zilio2020ALS}. Thus, the classifier needs to be robust towards these perturbations. 

Generating adversarial examples in text is challenging compared to computer vision tasks because of (i) discrete nature of input space and (ii) preservation of semantic coherence with original text. Initial works for attacking text models relied on introducing errors at the character level or manipulating words \cite{DBLP:journals/corr/abs-1804-07781} to generate adversarial examples. But due to grammatical disfluency, these seem very unnatural. Some rule-based synonym replacement strategies \cite{alzantot}, \cite{ren-etal-2019-generating} have lead to more natural looking examples. \cite{textfooler} proposed TextFooler, as a baseline to generate adversaries for text classification models. But, the adversarial examples created by TextFooler rely heavily on word-embedding based word similarity replacement technique, and not overall sentence semantics. Recently, \cite{Garg2020BAEBA} proposed BERT-MLM-based \cite{devlin-etal-2019-bert} word replacements to create adversaries to better fit the overall context.

Despite these advancements, there is much less attention towards making robust predictions in critical domains like biomedical, which comes with its unique challenges. \cite{Araujo2020OnAE} has proposed two types of rule-based adversarial attacks inspired by natural spelling errors and typos made by humans and synonym replacement in the biomedical domain. Some challenges include: 1) Biomedical named entities are usually \emph{multi-word phrases} such as \emph{colorectal adenoma}. During token replacement, we need the entire entity to be replaced, but the MLM model (token-level replacement) fails to generate correct synonym of entity fitting in the context. So, we need a BioNER+Entity Linker \cite{martins-etal-2019-joint}, \cite{mondal-etal-2019-medical} to link entity to ontology for generating correct synonyms. 2) Due to several variations of representing medical entities such as \emph{Type I Diabetes} could be expressed as \emph{'Type One Diabetes'}, we explore \emph{numeric entity expansion} strategies for generating adversaries. 3) Spelling variations (keyboard swap, modification). While we evaluate on two benchmark datasets, our method is general and is applicable for any biomedical classification datasets.

In this paper, we present \textbf{BBAEG (Biomedical BERT-based Adversarial Example Generation)}\footnote{https://github.com/Ishani-Mondal/BBAEG.git}, a novel black-box attack algorithm for biomedical text classification task leveraging both the BERT-MLM model for non-named entity replacements combined with NER linked synonyms for named entities to better fit the overall context. In addition to replacing words with synonyms, we explore the mechanism of generating adversarial examples using \emph{typographical variations and numeric entity modification}. Our BBAEG attack beats the existing baselines by a wide margin on both automatic and human evaluation across datasets and models. To the best of our knowledge, we are the first to introduce a novel algorithm for generating adversarial examples for biomedical text whose success attack is higher than the existing baselines like TextFooler and BAE \cite{Garg2020BAEBA}. \emph{The overall contributions of the paper} include: \textbf{1)} We explore several challenges of biomedical adversarial example generation. \textbf{2)} We propose BBAEG, a biomedical adversarial example generation technique for text classification combining the power of several perturbation techniques. \textbf{3)} We introduce 3 type of attacks for this purpose on two biomedical text classification datasets. \textbf{4)} Through human evaluation, we show that BBAEG yields adversarial examples with improved naturalness. 

\begin{algorithm}[!t]
  \SetAlgoLined
  \small
  \KwIn{$D$=[$w_1$, ... $w_l$], label = $y$, target classification model $M$}
  \KwOut{Adversarial example of $D$ = $D_{adv}$}
  \textbf{Initialization: $D_{adv}$ $\leftarrow$ D}, Tag the entities in $D$, Named entities are in $S_{NE}$ and the rest in $S_{NNE}$ \;
  Compute token importance $I_i$ $\forall$ $w_i$ $\in$ $D$\;
  \For{$i$ in descending order of $I_i$}{
    L = \{\} \;
    \uIf{($w_i$ in $S_{NE}$ and $(w_{i-t}..w_{i+t})$ is a NE)}{
       Syns = synonyms of NE\;
       \For{s $\in$ Syns}{
        $L[s]$ = $D_{adv[1:i-t-1]}$[s]$D_{adv[i+t+1:l]}$
       }
       \textbf{end for}\;
       }
       \ElseIf{($w_i$ in $S_{NNE}$)}{
        $D_{adv}$ = $D_{adv[1:i-1]}$[M]$D_{adv[i+1:l]}$\;
        $T$ = top-$K$ filtered and semantically similar tokens for $M$ $\in$ $D_{M}$\;
        \For{$t$ $\in$ $T$}{
            $L[t]$ = $D_{adv[1:i-1]}$[t]$D_{adv[i+1:l]}$
        }
        \textbf{end for}\;
        }
        \textbf{end if}\;
        \uIf{$\exists$ $t$ $\in$ $T$ such that $M(L[t])$ $\ne$ $y$}{
            \textbf{Return:} $D_{adv}$ $\leftarrow$ $L[t']$ where $M(L[t])$ $\ne$ $y$ and $L[t']$ has maximum similarity with $D$
        }
        \uElse{
            $N_1$ = Rotate $p$ characters in $w_i$ ($p$ $\leq$ $l$)\;
           $N_2$ = Random insertion of symbols before/end in $w_i$\;
            Noise = $N_1$ + $N_2$ \;
            \For{$t$ $\in$ $Noise$}{
                $L[t]$ = $D_{adv[1:i-1]}$[t]$D_{adv[i+1:l]}$
            }
            \textbf{end for}\;
            \uIf{$\exists$ $t$ $\in$ $T$ such that $M(L[t])$ $\ne$ $y$}{
            \textbf{Return:} $D_{adv}$ $\leftarrow$ $L[t']$ where $M(L[t])$ $\ne$ $y$ and $L[t']$ has maximum similarity with $D$
        }
            \uElseIf{$w_i$ contains numeric entity}{
                $t$ = Replace $w_i$ by $num2words$ \;
                $L[t]$ = $D_{adv[1:i-1]}$[t]$D_{adv[i+1:l]}$\;
                \textbf{Return:} $D_{adv}$ $\leftarrow$ $L[t]$ if $M(L[t])$ $\ne$ $y$
             }
             \uElse{
             \textbf{Return:} $D_{adv}$ $\leftarrow$ $L[t']$ where $L[t']$ causes max reduction in $y$ probability
             }
             \textbf{end if}\;
    }
    \textbf{end if}\;
  }
  \textbf{end for}\;
  \textbf{Return} $D_{adv}$ $\leftarrow$ None
  \caption{BBAEG Algorithm}
\end{algorithm}

\section{Methodology}
\textbf{Problem Definition:} Given a set of $n$ inputs
($D, Y$) = [($D_1$, $y_1$), . . .($D_n$, $y_n$)] and a trained
classifier $M$ : $D$ $\rightarrow$ $Y$, we assume the
soft-label black-box setting where the attacker can
only query the classifier for output probabilities on
a given input, and has no access to the model
parameters, gradients or training data. For an input of length $l$ consisting of words $w_i$, where 1 $\le$ $i$ $\le$ $l$,  $(D_i=[w_1, . . . , w_l], y)$, we want to generate an adversarial example $D_{adv}$ such that $M(D_{adv})$ $\ne$ $y$. We would like $D_{adv}$ to be grammatically correct, semantically similar to $D$ ($Sim$($D$, $D_{adv}$) $\ge$ $\alpha$), where $\alpha$ denotes the similarity threshold. \\

\noindent
\textbf{BBAEG Algorithm:} \\
Our proposed \textbf{BBAEG algorithm} consists of four steps:
\textbf{1)} Tagging the biomedical entities on $D$ and prepare two classes NE (named entities) and Non-NE (non-named entities) \textbf{2)} Ranking the important words for perturbation \textbf{3)} Choosing perturbation schemes \textbf{4)} Final adversaries generation. \\

\noindent
\underline{\textbf{1) Named Entity Tagging:}}
For each input instance $D_i$ (Line 1 in Algorithm), we apply sciSpacy\footnote{https://allenai.github.io/scispacy/} with \emph{en-ner-bc5cdr-md} to extract biomedical named entities (drugs and diseases), followed by its Entity Linker (Drugs to DrugBank \cite{drugbank}, Disease to MESH\footnote{https://meshb.nlm.nih.gov/})). After linking the NE to respective ontologies, we use pyMeshSim\footnote{https://github.com/luozhhub/pyMeSHSim} (for disease) and DrugBank (for drugs) to obtain synonyms. In each $D_i$ of size $l$ ($w_1, w_2,... [w_i...w_{i+2}],...w_l$), multi-word expressions ($w_i...w_{i+2}$) are named entities. We put them in Named Entities Set $(S_{NE})$ and other words in non-Named Entity set $(S_{NNE})$.\\

\noindent
\underline{\textbf{2) Ranking of important words:}} We estimate token importance $I_i$ of each $w_i$ $\in$ $D$, by deleting $w_i$ from $D$ and computing the decrease in probability of predicting the correct label $y$ (Line 2), similar to \cite{textfooler}. Thus, we receive a set for each token which contains the tokens in decreasing order of their importance. \\

\noindent
\underline{\textbf{3) Choosing perturbation schemes:}} Consider the input $D_i$, we describe a \textit{sieve-based approach} of perturbing $D_i$. Sieves are ordered by precision, with the most precise sieve appearing first. \\

\noindent
\underline{\textbf{Sieve 1 :}} In the first sieve, we propose to alter the synonyms of the tokens in $S_{NE}$ (Line 5-9) using Ontology linking and the words in $S_{NNE}$ (Line 10-15) using BERT-MLM predicted tokens. This stems from the fact that synonym replacement of the non-named entities using \textbf{BERT-MLM} generates reasonable predictions considering the surrounding context \cite{Garg2020BAEBA}. If the token is a part of $S_{NE}$, replace them with the domain-specific synonyms one by one, but if the token is part of $S_{NNE}$, then replace those words by the top-$K$ BERT-MLM predictions. To achieve high semantic similarity with the original text, we filter the set of top $K$ tokens ($K$ is a pre-defined constant) (Line 12) predicted by BERT-MLM for the masked token, using a Sentence-Transformer \cite{sbert} based sentence similarity scorer. Additionally, we filter out predicted tokens that do not belong to the same part of speech as original token.  If this sieve generates adversaries for $D_i$, then $D_{adv}$ is being returned.\\

\noindent
\underline{\textbf{Sieve 2:}} (Line 20-28) If the first sieve does not generate adversary, we introduce two typographical noise in the input 1) \textbf{Spelling Noise-N1:} Rotating random $p$ characters (Line 20) 2) \textbf{Spelling Noise-N2:} insertion of symbols to the beginning or end (Line 21). If this sieve generates adversaries for $D_i$, then $D_{adv}$ is being returned. \\

\noindent
\underline{\textbf{Sieve 3:}} (Line 29-31) If Sieve 2 does not generate adversary, we replace the numeric entities by expanding the numeric digit. For example: \emph{PMD1} can be rewritten as {PMD One}, \emph{Covid19} as \emph{Covid nineteen}. If this sieve generates adversaries for $D_i$, then $D_{adv}$ is being returned. \\

\noindent
\underline{\textbf{4) Final adversaries generation:}} For each of the three sieves, among all the winning adversaries, the one which is the most similar to original text as measured by \cite{sbert} is returned. If the sieves do not generate adversaries, we return the perturbed example which causes maximum reduction in the probability of output.

\begin{table*}[!t]
\small
\begin{tabular}{l|ccc|ccc}
\hline
 && \textbf{Twitter ADE Corpus} &&& \textbf{ADE} \\
 & $\textbf{Before-attack}$ & $\textbf{After-attack}$ & $\textbf{\%}$ & $\textbf{Before-attack}$ & $\textbf{After-attack}$  &\textbf{\%} \\
\hline
\textbf{HAN-TF} & 0.80 & \textbf{0.33} & 0.10 & 0.83 & 0.46 & 0.09 \\
\textbf{HAN-BAE} & 0.80 & 0.35 & 0.08 & 0.83 & 0.43 & \textbf{0.06} \\
\textbf{HAN-Ours} & 0.80 & 0.36 & \textbf{0.05} & 0.83 & \textbf{0.31} & 0.11 \\
\hline

\textbf{BERT-base-TF} & 0.83 & 0.52 & \textbf{0.12} & 0.85 & 0.59 & \textbf{0.11}\\
\textbf{BERT-base-BAE} & 0.83 & 0.50 & 0.16 & 0.85 & 0.60 & 0.15 \\
\textbf{BERT-base-BBAEG} & 0.83 & \textbf{0.44} & \textbf{0.12} & 0.85 & \textbf{0.54} & 0.13\\
\hline

\textbf{RoBERTa-base-TF} & 0.82 & 0.66 & 0.26 & 0.86 & 0.75 & 0.28\\
\textbf{RoBERTa-base-BAE} & 0.82 & 0.63 & 0.23 & 0.86 & 0.74 & 0.24 \\
\textbf{RoBERTa-base-BBAEG} & 0.82 & \textbf{0.57} & \textbf{0.19} & 0.86 & \textbf{0.70} & \textbf{0.23}\\
\hline

\textbf{SciBERT-TF} & 0.85 & 0.45 & 0.11 & 0.88 & 0.53 & 0.13\\
\textbf{SciBERT-BAE} & 0.85 & 0.43 & 0.11 & 0.88 & 0.56 & 0.11\\
\textbf{SciBERT-BBAEG} & 0.85 & \textbf{0.38} & \textbf{0.10} & 0.88 & \textbf{0.50} & \textbf{0.08} \\
\hline

\textbf{BioBERT-TF} & 0.86 & 0.51 & 0.18 & 0.87 & 0.51 & 0.09 \\
\textbf{BioBERT-BAE} & 0.86 & 0.48 & \textbf{0.13} & 0.87 & 0.48 & 0.13\\
\textbf{BioBERT-BBAEG} & 0.86 & \textbf{0.37} & \textbf{0.13} & 0.87 & \textbf{0.45} & \textbf{0.07}\\
\hline

\textbf{ClinicalBERT-TF} & 0.81 & 0.47 & 0.17 & 0.81 & 0.54 & \textbf{0.15} \\
\textbf{ClinicalBERT-BAE} & 0.81 & 0.48 & \textbf{0.16} & 0.81 & 0.58 & 0.22 \\
\textbf{ClinicalBERT-BBAEG} & 0.81 & \textbf{0.46} & 0.17 & 0.81 & \textbf{0.50} & 0.19 \\
\hline

\end{tabular}
\caption{Before-attack and after-attack accuracies of the models along with the \% of perturbed words in the input space. Best attack and least \% of perturbations are shown in \textbf{bold} for each dataset.}
\end{table*}

\begin{table*}[t]
\tiny
\caption{shows the adversaries generated by BBAEG on handpicked examples from test set of ADE corpus. The different adversaries generated by baselines and BBAEG are shown. Also, the adversaries generated using different ablation of sieves [Spellings in \textcolor{blue} {Blue} and Number in \textcolor{green} {green}, synonyms by attack algorithms in \textcolor{red}{red}] are shown.}
\label{tab-0}
\centering
\begin{tabular}{c}
\textbf{Adverse Drug Event (ADE) Corpus (Adversaries : ADE Present $\rightarrow$ ADE Not present)} \\
\end{tabular}
\begin{tabular}{l|l}
\hline
\textbf{Original:} & Successful challenge with \textbf{clozapine} in a history of \textbf{pulmonary eosinophilia} ailment. \\
\textbf{BAE (Using BERT-MLM):} & Successful challenge with \textcolor{red}{hydrochloride} in a history of pulmonary \textcolor{red}{disease} ailment.\\
\textbf{BBAEG (Best Combination):} & Successful challenge with \textcolor{red}{clozapinum} in a history of \textcolor{red}{Loeffler Syndrome} ailment. \\
\hline
\textbf{Original:} & A 21-year-old patient developed rhabdomyolysis during 19th week of treatment with \textbf{clozapine}  for \textbf{schizophrenia}.\\
\textbf{BBAEG (Spelling Noise-N2):} & A 21-year-old patient developed rhabdomyolysis during 19th week of treatment with \textcolor{blue}{inoclozapine} for \textcolor{blue}{cdschizophrenia}.\\
\textbf{BBAEG (Spelling Noise-N1):} & A 21-year-old patient developed rhabdomyolysis during 19th week of treatment with \textcolor{blue}{clpazoine} for  \textcolor{blue}{schizoerhpnia}.\\
\textbf{BBAEG (Synonyms):} & A 21-year-old patient developed rhabdomyolysis during 19th week of treatment with \textcolor{red}{Clozapinum} for  \textcolor{red}{dementia Praecox}.\\
\textbf{BBAEG (Number Replacement):} & A \textcolor{green}{twenty-one}-year-old patient developed rhabdomyolysis during \textcolor{green}{nineteen}th week of treatment with clozapine  for schizophrenia.\\
\hline
\end{tabular}
\end{table*}

\section{Experimental setup}
\textbf{Datasets and Experimental Details:} We evaluate BBAEG on two different biomedical text classification datasets: 1) Adverse Drug Event (ADE) Detection \cite{GURULINGAPPA2012885} and 2) Twitter ADE dataset \cite{rosenthal-etal-2017-semeval} for the task of classifying whether the sentence contains mention of ADE (binary).

We use 6 classification models as $M$: Hierarchical Attention Model \cite{han}, BERT \cite{devlin-etal-2019-bert}, RoBERTa \cite{roberta}, BioBERT \cite{biobert}, Clinical-BERT \cite{clinicalbert}, SciBERT \cite{scibert}. We fine-tune these models on the training data (of each corpus) using Adam Optimizer \cite{Kingma2015AdamAM} with learning rate of 0.00002, 10 epochs and perform adversarial attack on the test data. For the BBAEG non-NER synonym attacks, we use BERT-base-uncased MLM to predict the masked tokens. We consider top $K$=10 synonyms from the BERT-MLM predictions and set threshold $\alpha$ of 0.75 for cosine similarity between \cite{sbert} embeddings of the adversarial and input text, we set $p$=2 characters for rotation to introduce noise in input. For more details refer to the appendix.

\section{Results}
\textbf{Automatic Evaluation Results:} 
We examine the success of adversarial attack using two criteria: \textbf{(1) Performance Drop} (Adrop): Difference between original (accuracy on original test set) and after-attack accuracy (accuracy on the perturbed test set) \textbf{(2) Perturbation of input (\%):} Percentage of perturbed words in adversary generated. Success of attack is directly and indirectly proportional with criteria 1 and 2 respectively.\\

\noindent
\underline{\textbf{Effectiveness:}} Table 1 shows the results of BBAEG attack on two datasets across all the models. During our experiments with HAN (general deep learning model), we observe that the attack is the most successful compared to BERT-variants, RoBERTa and the existing baselines, in terms of both the criteria (1 and 2). Also, using BioBERT  and Sci-BERT (35-45\% and 40-50\% accuracy drop respectively), the attack is the most successful. This stems from the fact that the vocabularies used in the datasets have already been explored during pre-training by the contextual embeddings, thus more sensitive towards small perturbations. Moreover, it has been clearly observed that unlike BERT and HAN, RoBERTa is very less susceptible to adversarial attacks (10-20\% accuracy drop), perturbing 20-25\% words in the input space. We also observe that BERT-MLM-based synonym replacement techniques for non-NER, combined with \emph{multi-word NER} synonym replacement using entity linking outperforms TextFooler(TF) and BAE-based approaches in terms of accuracy drop.

\begin{table*}[h]
\small
\centering
\begin{tabular}{l|c|c}
\hline
 & \textbf{Twitter ADE} & \textbf{ADE} \\
& Accuracy Drop (Semantic Similarity) & Accuracy Drop (Semantic Similarity) \\
\hline
\textbf{BioBERT-BBAEG (best variation)} 
 & 0.43 (0.893) & 0.42 (0.906) \\

- w/o Synonym Replacement (S1)
 & 0.39 (0.899) & 0.40 (0.919) \\
 
- w/o Spelling Noise N1 (S2-1)
  & 0.37 (0.901) & 0.35 (0.912) \\
 
- w/o Spelling Noise N2 (S2-2)
 & 0.34 (0.913) & 0.31 (0.891)) \\
 
- w/o Number Replacement (S3) & 0.30 (0.920) & 0.27 (0.915) \\
 \hline
 \textbf{SciBERT-BBAEG (best variation)} 
 & 0.45 (0.879)  & 0.38 (0.881) \\

- w/o Synonym Replacement (S1)
 & 0.42 (0.901) & 0.35 (0.912) \\
 
 - w/o Spelling Noise N1 (S2-1)
  & 0.39 (0.915) & 0.36 (0.901)  \\
 
  - w/o Spelling Noise N2 (S2-2)
 & 0.31 (0.891) & 0.31 (0.847) \\

- w/o Number Replacement (S3) & 0.32 (0.911)
 & 0.36 (0.903)  \\
 \hline
\caption{Ablation analysis of the sieves (S1-S3) on accuracy drop and average semantic similarities between adversaries  and original text.}
\end{tabular}
\end{table*}

\begin{table}[!t]
\tiny
\centering
\begin{tabular}{l|cc|cc}
 && \textbf{Twitter ADE} && \textbf{ADE} \\
\hline
 & Accuracy & Naturalness & Accuracy & Naturalness \\
\hline
\textbf{TextFooler (TF)} 
 & 0.85 & 3.78 & 0.78 & 3.55 \\

\textbf{BAE Algorithm} 
 & 0.88 & 3.95 & 0.84 & 3.89 \\
 
 \hline
\textbf{BBAEG (Our Method)} & \textbf{0.94} & \textbf{4.23} & \textbf{0.90} & \textbf{4.56} \\

\hline
\end{tabular}
\caption{Human Evaluation on both the datasets.}
\end{table}

\noindent
\underline{\textbf{Ablation Analysis:}} In Table 3, we perform an ablation analysis on the different perturbation schemes and the effect of the attack using each of the sieves by making use of two fine-tuned contextual embedding model as the target model for ADE classification. \emph{Synonym replacement (S1)} (average 35\% accuracy drop) and \emph{character rotation (S2-1)} (average 38\% accuracy drop) seems to be the most promising approach for success attacks on biomedical text classification. Moreover, we conduct a deeper analysis to gain an insight of how much the synonyms of \emph{NER vs Non-NER entities} contribute towards prediction change. We have found that the multi-word NERs during replacement generates natural-looking examples (compared to MLM-based entity replacement such as \emph{pulmonary eosinophillia} is replaced by \emph{Loeffler Syndrome} (for BBAEG) by normalizing to MESH vocabulary, while replaced by \emph{disease} in BAE predictions as shown in Table 2 and they seem very unnatural. This proves that high semantic similarity does not always ensure generation of proper grammatical adversaries. \\

\noindent
\underline{\textbf{Human Evaluation:}} Apart from automatic evaluation, we also perform human evaluation of our BBAEG attacks on the BERT classifier. We perform similar kind of human evaluation by two biomedical domain-experts on randomly selected 100 generated adversarial examples (from each of the different attack algorithms) on each of the two datasets. For each sample, 50 annotations were collected. Similar setup was performed by \cite{Garg2020BAEBA} during evaluation. The main two criteria for evaluation of the perturbed samples are as follows: \\

\noindent
\textbf{1) Naturalness :} How much the adversaries generated is semantically similar to the original text content, preserving grammatical correctness on Likert Scale (1-5)? To evaluate the naturalness of the adversarial examples, we first present the annotators with 50 different set of original data
samples to understand data distribution. \\

\noindent
\textbf{2) Accuracy of  generated instances:} on the binary classification of presence of \emph{Adverse Drug Reaction (ADR)} on the adversarial examples. We enumerate the average scores of two annotators (for TextFooler (TF), BAE and our BBAEG) and present those in Table 4.

During ablation analysis, we observe that the synonym replaced perturbed samples looked more natural to the human evaluators compared to the spelling perturbed samples and number replaced entities. When considered jointly, the number replaced and synonym replaced samples seemed more natural to the annotators compared to spelling perturbed samples. This arises due to the fact that the number replaced entities when thrown to the annotators they could easily interpret the meaning correctly when given in combination with the original sample. For instance, in the examples shown in table 2, the number replaced samples (\emph{21-year old} $\rightarrow$ \emph{twenty-one-year old}) look more natural and easily interpretable compared to spelling perturbed samples (\emph{clozapine} $\rightarrow$ \emph{clpazoine}).

\section{Conclusion and Future Work}
In this paper, we propose a new technique for generating adversarial examples combining contextual perturbations based on BERT-MLM, synonym replacement of biomedical entities, typographical errors and numeric entity expansion. We explore several classification models to demonstrate the efficacy of our method. Experiments conducted on two benchmark biomedical datasets demonstrate the strength and effectiveness of our attack. As a future work, we would like to explore more about retraining the models with the perturbed samples in order to improve model robustness.

\section*{Acknowledgement}
The author would like to thank the annotators for hard work, and also the anonymous reviewers for their insightful comments and feedback.

\bibliography{anthology,naacl2021}

\begin{thebibliography}{23}
\expandafter\ifx\csname natexlab\endcsname\relax\def\natexlab#1{#1}\fi

\bibitem[{Alzantot et~al.(2018)Alzantot, Sharma, Elgohary, Ho, Srivastava, and
  Chang}]{alzantot}
Moustafa Alzantot, Yash Sharma, Ahmed Elgohary, Bo-Jhang Ho, Mani Srivastava,
  and Kai-Wei Chang. 2018.
\newblock \href {https://doi.org/10.18653/v1/D18-1316} {Generating natural
  language adversarial examples}.
\newblock In \emph{Proceedings of the 2018 Conference on Empirical Methods in
  Natural Language Processing}, pages 2890--2896, Brussels, Belgium.
  Association for Computational Linguistics.

\bibitem[{Araujo et~al.(2020)Araujo, Carvallo, Aspillaga, and
  Parra}]{Araujo2020OnAE}
Vladimir Araujo, Andres Carvallo, C.~Aspillaga, and Denis Parra. 2020.
\newblock On adversarial examples for biomedical nlp tasks.
\newblock \emph{ArXiv}, abs/2004.11157.

\bibitem[{Beltagy et~al.(2019)Beltagy, Cohan, and Lo}]{scibert}
Iz~Beltagy, Arman Cohan, and Kyle Lo. 2019.
\newblock \href {http://arxiv.org/abs/1903.10676} {Scibert: Pretrained
  contextualized embeddings for scientific text}.
\newblock \emph{CoRR}, abs/1903.10676.

\bibitem[{Campillos~Llanos et~al.(2017)Campillos~Llanos, Rosset, and
  Zweigenbaum}]{campillos-llanos-etal-2017-automatic}
Leonardo Campillos~Llanos, Sophie Rosset, and Pierre Zweigenbaum. 2017.
\newblock \href {https://doi.org/10.18653/v1/W17-2343} {Automatic
  classification of doctor-patient questions for a virtual patient record query
  task}.
\newblock In \emph{{B}io{NLP} 2017}, pages 333--341, Vancouver, Canada,.
  Association for Computational Linguistics.

\bibitem[{Devlin et~al.(2019)Devlin, Chang, Lee, and
  Toutanova}]{devlin-etal-2019-bert}
Jacob Devlin, Ming-Wei Chang, Kenton Lee, and Kristina Toutanova. 2019.
\newblock \href {https://doi.org/10.18653/v1/N19-1423} {{BERT}: Pre-training of
  deep bidirectional transformers for language understanding}.
\newblock In \emph{Proceedings of the 2019 Conference of the North {A}merican
  Chapter of the Association for Computational Linguistics: Human Language
  Technologies, Volume 1 (Long and Short Papers)}, pages 4171--4186,
  Minneapolis, Minnesota. Association for Computational Linguistics.

\bibitem[{Feng et~al.(2018)Feng, Wallace, Iyyer, Rodriguez, II, and
  Boyd{-}Graber}]{DBLP:journals/corr/abs-1804-07781}
Shi Feng, Eric Wallace, Mohit Iyyer, Pedro Rodriguez, Alvin~Grissom II, and
  Jordan~L. Boyd{-}Graber. 2018.
\newblock \href {http://arxiv.org/abs/1804.07781} {Right answer for the wrong
  reason: Discovery and mitigation}.
\newblock \emph{CoRR}, abs/1804.07781.

\bibitem[{Garg and Ramakrishnan(2020)}]{Garg2020BAEBA}
Siddhant Garg and Goutham Ramakrishnan. 2020.
\newblock Bae: Bert-based adversarial examples for text classification.
\newblock In \emph{EMNLP}.

\bibitem[{Gurulingappa et~al.(2012)Gurulingappa, Rajput, Roberts, Fluck,
  Hofmann-Apitius, and Toldo}]{GURULINGAPPA2012885}
Harsha Gurulingappa, Abdul~Mateen Rajput, Angus Roberts, Juliane Fluck, Martin
  Hofmann-Apitius, and Luca Toldo. 2012.
\newblock \href {https://doi.org/https://doi.org/10.1016/j.jbi.2012.04.008}
  {Development of a benchmark corpus to support the automatic extraction of
  drug-related adverse effects from medical case reports}.
\newblock \emph{Journal of Biomedical Informatics}, 45(5):885 -- 892.
\newblock Text Mining and Natural Language Processing in Pharmacogenomics.

\bibitem[{Huang et~al.(2019)Huang, Altosaar, and Ranganath}]{clinicalbert}
Kexin Huang, Jaan Altosaar, and Rajesh Ranganath. 2019.
\newblock \href {http://arxiv.org/abs/1904.05342} {Clinicalbert: Modeling
  clinical notes and predicting hospital readmission}.
\newblock \emph{CoRR}, abs/1904.05342.

\bibitem[{Jin et~al.(2019)Jin, Jin, Zhou, and Szolovits}]{textfooler}
Di~Jin, Zhijing Jin, Joey~Tianyi Zhou, and Peter Szolovits. 2019.
\newblock \href {http://arxiv.org/abs/1907.11932} {Is {BERT} really robust?
  natural language attack on text classification and entailment}.
\newblock \emph{CoRR}, abs/1907.11932.

\bibitem[{Kazi and Kahanda(2019)}]{kazi-kahanda-2019-automatically}
Nazmul Kazi and Indika Kahanda. 2019.
\newblock \href {https://doi.org/10.18653/v1/W19-1918} {Automatically
  generating psychiatric case notes from digital transcripts of doctor-patient
  conversations}.
\newblock In \emph{Proceedings of the 2nd Clinical Natural Language Processing
  Workshop}, pages 140--148, Minneapolis, Minnesota, USA. Association for
  Computational Linguistics.

\bibitem[{Kingma and Ba(2015)}]{Kingma2015AdamAM}
Diederik~P. Kingma and Jimmy Ba. 2015.
\newblock Adam: A method for stochastic optimization.
\newblock \emph{CoRR}, abs/1412.6980.

\bibitem[{Lee et~al.(2019)Lee, Yoon, Kim, Kim, Kim, So, and Kang}]{biobert}
Jinhyuk Lee, Wonjin Yoon, Sungdong Kim, Donghyeon Kim, Sunkyu Kim, Chan~Ho So,
  and Jaewoo Kang. 2019.
\newblock \href {http://arxiv.org/abs/1901.08746} {Biobert: a pre-trained
  biomedical language representation model for biomedical text mining}.
\newblock \emph{CoRR}, abs/1901.08746.

\bibitem[{Liu et~al.(2019)Liu, Ott, Goyal, Du, Joshi, Chen, Levy, Lewis,
  Zettlemoyer, and Stoyanov}]{roberta}
Yinhan Liu, Myle Ott, Naman Goyal, Jingfei Du, Mandar Joshi, Danqi Chen, Omer
  Levy, Mike Lewis, Luke Zettlemoyer, and Veselin Stoyanov. 2019.
\newblock \href {http://arxiv.org/abs/1907.11692} {Roberta: {A} robustly
  optimized {BERT} pretraining approach}.
\newblock \emph{CoRR}, abs/1907.11692.

\bibitem[{Martins et~al.(2019)Martins, Marinho, and
  Martins}]{martins-etal-2019-joint}
Pedro~Henrique Martins, Zita Marinho, and Andr{\'e} F.~T. Martins. 2019.
\newblock \href {https://doi.org/10.18653/v1/P19-2026} {Joint learning of named
  entity recognition and entity linking}.
\newblock In \emph{Proceedings of the 57th Annual Meeting of the Association
  for Computational Linguistics: Student Research Workshop}, pages 190--196,
  Florence, Italy. Association for Computational Linguistics.

\bibitem[{Mondal et~al.(2019)Mondal, Purkayastha, Sarkar, Goyal, Pillai,
  Bhattacharyya, and Gattu}]{mondal-etal-2019-medical}
Ishani Mondal, Sukannya Purkayastha, Sudeshna Sarkar, Pawan Goyal, Jitesh
  Pillai, Amitava Bhattacharyya, and Mahanandeeshwar Gattu. 2019.
\newblock \href {https://doi.org/10.18653/v1/W19-1912} {Medical entity linking
  using triplet network}.
\newblock In \emph{Proceedings of the 2nd Clinical Natural Language Processing
  Workshop}, pages 95--100, Minneapolis, Minnesota, USA. Association for
  Computational Linguistics.

\bibitem[{Reimers and Gurevych(2019)}]{sbert}
Nils Reimers and Iryna Gurevych. 2019.
\newblock \href {http://arxiv.org/abs/1908.10084} {Sentence-bert: Sentence
  embeddings using siamese bert-networks}.
\newblock \emph{CoRR}, abs/1908.10084.

\bibitem[{Ren et~al.(2019)Ren, Deng, He, and Che}]{ren-etal-2019-generating}
Shuhuai Ren, Yihe Deng, Kun He, and Wanxiang Che. 2019.
\newblock \href {https://doi.org/10.18653/v1/P19-1103} {Generating natural
  language adversarial examples through probability weighted word saliency}.
\newblock In \emph{Proceedings of the 57th Annual Meeting of the Association
  for Computational Linguistics}, pages 1085--1097, Florence, Italy.
  Association for Computational Linguistics.

\bibitem[{Rosenthal et~al.(2017)Rosenthal, Farra, and
  Nakov}]{rosenthal-etal-2017-semeval}
Sara Rosenthal, Noura Farra, and Preslav Nakov. 2017.
\newblock \href {https://doi.org/10.18653/v1/S17-2088} {{S}em{E}val-2017 task
  4: Sentiment analysis in {T}witter}.
\newblock In \emph{Proceedings of the 11th International Workshop on Semantic
  Evaluation ({S}em{E}val-2017)}, pages 502--518, Vancouver, Canada.
  Association for Computational Linguistics.

\bibitem[{Wishart et~al.(2017)Wishart, Djoumbou, Guo, Lo, Marcu, Grant, Sajed,
  Johnson, Li, Sayeeda, Assempour, Iynkkaran, Liu, Maciejewski, Gale, Wilson,
  Chin, Cummings, Le, and Wilson}]{drugbank}
David Wishart, Yannick Djoumbou, An~Chi Guo, Elvis Lo, Ana Marcu, Jason Grant,
  Tanvir Sajed, Daniel Johnson, Carin Li, Zinat Sayeeda, Nazanin Assempour,
  Ithayavani Iynkkaran, Yifeng Liu, Adam Maciejewski, Nicola Gale, Alex Wilson,
  Lucy Chin, Ryan Cummings, Diana Le, and Michael Wilson. 2017.
\newblock \href {https://doi.org/10.1093/nar/gkx1037} {Drugbank 5.0: A major
  update to the drugbank database for 2018}.
\newblock \emph{Nucleic acids research}, 46.

\bibitem[{Yang et~al.(2016)Yang, Yang, Dyer, He, Smola, and Hovy}]{han}
Zichao Yang, Diyi Yang, Chris Dyer, Xiaodong He, Alex Smola, and Eduard Hovy.
  2016.
\newblock \href {https://doi.org/10.18653/v1/N16-1174} {Hierarchical attention
  networks for document classification}.
\newblock pages 1480--1489.

\bibitem[{Zhang et~al.(2020)Zhang, Jiang, Zhang, Liu, Cao, Liu, Liu, and
  Zhao}]{Zhang2020MIEAM}
Yuanzhe Zhang, Z.~Jiang, T.~Zhang, Shiwan Liu, Jiarun Cao, Kang Liu, Shengping
  Liu, and Jun Zhao. 2020.
\newblock Mie: A medical information extractor towards medical dialogues.
\newblock In \emph{ACL}.

\bibitem[{Zilio et~al.(2020)Zilio, Paraguassu, Hercules, Ponomarenko,
  Berwanger, and Finatto}]{Zilio2020ALS}
L.~Zilio, Liana~Braga Paraguassu, Luis Antonio~Leiva Hercules, G.~Ponomarenko,
  Laura Berwanger, and Maria Jos{\'e}~Bocorny Finatto. 2020.
\newblock A lexical simplification tool for promoting health literacy.
\newblock In \emph{READI}.

\end{thebibliography}
\bibliographystyle{acl_natbib}

\end{document}